\documentclass[11pt,a4paper]{article}
\usepackage[hyperref]{aacl-ijcnlp2020}
\usepackage{times}
\usepackage{latexsym}

\usepackage{microtype}

\aclfinalcopy 
\def\aclpaperid{312} 

\usepackage{microtype}
\usepackage{latexsym} 
\usepackage{graphicx}
\usepackage{wrapfig}
\usepackage{amssymb}
\usepackage{color,xcolor,colortbl}
\usepackage{enumitem}
\usepackage{soul}
\usepackage{amsmath}
\usepackage{url}
\usepackage{algorithm}
\usepackage{algorithmic}
\usepackage{booktabs}
\usepackage{bm,bbm}
\usepackage{mathtools}
\usepackage{array}
\usepackage{multirow}
\usepackage{subcaption}
\usepackage{pbox}
\usepackage{pifont}
\usepackage{arydshln}
\usepackage{dblfloatfix}
\usepackage{relsize}

\newcommand\numberthis{\addtocounter{equation}{1}\tag{\theequation}}

\definecolor{darkblue}{rgb}{0.0, 0.0, 0.55}
\setulcolor{blue} 
\usepackage{soul}
\newcommand{\hlc}[2][yellow]{{\sethlcolor{#1}\hl{#2}}}
\definecolor{d}{HTML}{c6dbef}
\definecolor{l}{HTML}{e5f5e0}

\title{A Cascade Approach to Neural Abstractive Summarization with \\Content Selection and Fusion}

\author{Logan Lebanoff,$^{\spadesuit}$ \, Franck Dernoncourt,$^{\diamondsuit}$ \, Doo Soon Kim,$^{\diamondsuit}$ \, Walter Chang,$^{\diamondsuit}$ \, Fei Liu$^{\spadesuit}$\\
  $^{\spadesuit}$Computer Science Department, University of Central Florida, Orlando, FL\\
  $^{\diamondsuit}$Adobe Research, San Jose, CA\\
  {\tt loganlebanoff@knights.ucf.edu \{dernonco,dkim,wachang\}@adobe.com}\\ 
  {\tt feiliu@cs.ucf.edu} \\}
  
\date{}

\begin{document}
\maketitle

\begin{abstract}

We present an empirical study in favor of a cascade architecture to neural text summarization.
Summarization practices vary widely but few other than news summarization can provide a sufficient amount of training data enough to meet the requirement of end-to-end neural abstractive systems which perform content selection and surface realization jointly to generate abstracts. 
Such systems also pose a challenge to summarization evaluation, as they force content selection to be evaluated along with text generation, yet evaluation of the latter remains an unsolved problem.
In this paper, we present empirical results showing that the performance of a cascaded pipeline that separately identifies important content pieces and stitches them together into a coherent text is comparable to or outranks that of end-to-end systems, whereas a pipeline architecture allows for flexible content selection.
We finally discuss how we can take advantage of a cascaded pipeline in neural text summarization and shed light on important directions for future research.

\end{abstract}
 
\section{Introduction}

There is a variety of successful summarization applications but few can afford to have a large number of annotated examples that are sufficient to meet the requirement of end-to-end neural abstractive summarization.
Examples range from summarizing radiology reports~\cite{jing-etal-2019-show,zhang-etal-2020-optimizing} to congressional bills~\cite{kornilova-eidelman-2019-billsum} and meeting conversations~\cite{mehdad-etal-2013-abstractive,li-etal-2019-keep,koay-etal-2020-how}.
The lack of annotated resources suggests that end-to-end systems may not be a ``one-size-fits-all'' solution to neural text summarization.
There is an increasing need to develop cascaded architectures to allow for customized content selectors to be combined with general-purpose neural text generators to realize the full potential of neural abstractive summarization.

We advocate for explicit content selection as it allows for a rigorous evaluation and visualization of intermediate  results of such a module, rather than associating it with text generation. 
Existing neural abstractive systems can perform content selection implicitly using end-to-end models~\cite{see_get_2017,celikyilmaz-etal-2018-deep,Raffel:2019,lewis-etal-2020-bart}, or more explicitly, with an external module to select important sentences or words to aid generation~\cite{tan-etal-2017-abstractive,gehrmann-etal-2018-bottom,chen-bansal-2018-fast,kryscinski-etal-2018-improving,hsu-etal-2018-unified,lebanoff-etal-2018-adapting,lebanoff-etal-2019-scoring,liu-lapata-2019-hierarchical}.
However, content selection concerns not only the selection of important segments from a document, but also the cohesiveness of selected segments and the amount of text to be selected in order for a neural text generator to produce a summary.

\begin{figure*}
\centering
\includegraphics[width=5.5in]{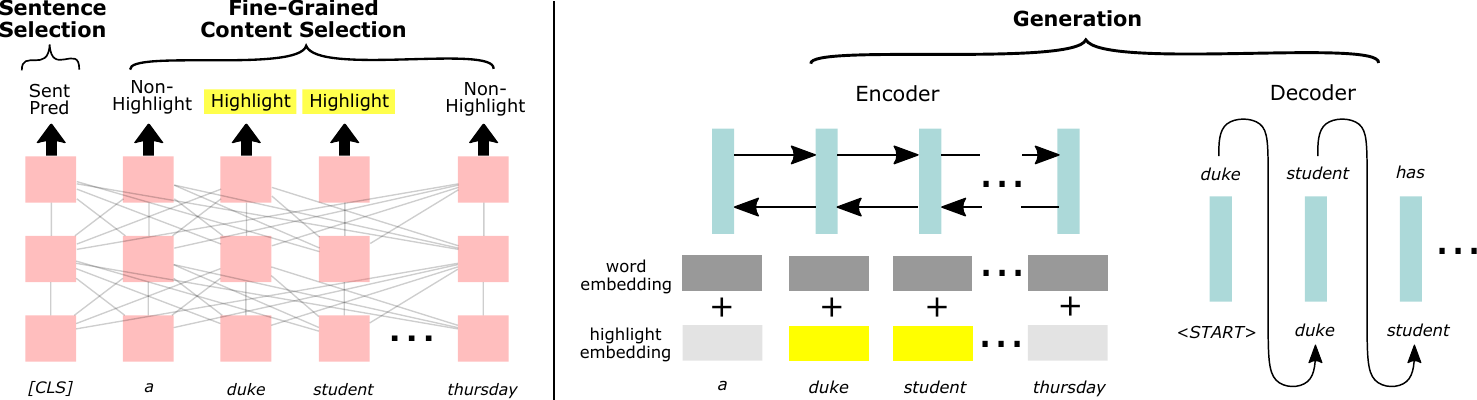}
\caption{
Model architecture.
We divide the task between two main components: the first component performs sentence selection and fine-grained content selection, which are posed as a classification problem and a sequence-tagging problem, respectively. 
The second component receives the first component's outputs as supplementary information to generate the summary.
A cascade architecture provides the necessary flexibility to separate content selection from surface realization in abstractive summarization.
}
\label{fig:model}
\vspace{-0.1in}
\end{figure*}

In this paper, we aim to investigate the feasibility of a cascade approach to neural text summarization.
We explore a constrained summarization task, where an abstract is created one sentence at a time through a cascaded pipeline.
Our pipeline architecture chooses one or two sentences from the source document, then highlights their summary-worthy segments and uses those as a basis for composing a summary sentence.
When a pair of sentences are selected, it is important to ensure that they are \emph{fusible}---there exists cohesive devices that tie the two sentences together into a coherent text---to avoid generating nonsensical outputs~\cite{geva-etal-2019-discofuse,lebanoff-etal-2020-understanding}. 
Highlighting sentence segments allows us to perform fine-grained content selection that guides the neural text generator to stitch selected segments into a coherent sentence. 
The contributions of this work are summarized as follows.
\begin{itemize}[topsep=5pt,itemsep=0pt]

\item We present an empirical study in favor of a cascade architecture for neural text summarization. Our cascaded pipeline chooses one or two sentences from the document and highlights their important segments; these segments are passed to a neural generator to produce a summary sentence.

\item Our quantitative results show that the performance of a cascaded pipeline is comparable to or outranks that of end-to-end systems, with added benefit of flexible content selection. We discuss how we can take advantage of a cascade architecture and shed light on important directions for future research.\footnote{Our code is publicly available at {\url{https://github.com/ucfnlp/cascaded-summ}}}

\end{itemize}

\section{A Cascade Approach} 
\label{sec:framework}

Our cascaded summarization approach focuses on shallow abstraction. 
It makes use of text transformations such as sentence shortening, paraphrasing and fusion~\cite{jing-mckeown-2000-cut} and is in contrast to deep abstraction, where a full semantic analysis of the document is often required.
A shallow approach helps produce abstracts that convey important information while, crucially, remaining faithful to the original.
In what follows, we describe our approach to select single sentences and sentence pairs from the document, highlight summary-worthy segments and perform summary generation conditioned on highlights.

\vspace{0.08in}
\noindent\textbf{Selection of Singletons and Pairs}\quad
Our approach iteratively selects one or two sentences from the input document; they serve as the basis for composing a single summary sentence.
Previous research suggests that 60-85\% of human-written summary sentences are created by shortening a single sentence or merging a pair of sentences~\cite{lebanoff-etal-2019-scoring}. 
We adopt this setting and present a coarse-to-fine strategy for content selection. 
Our strategy begins with selecting sentence singletons and pairs, followed by highlighting important segments of the sentences. 
Importantly, the strategy allows us to control which segments will be combined into a summary sentence---``compatible'' segments come from either a single document sentence or a pair of \emph{fusible} sentences.
In contrast, when all important segments of the document are provided to a neural generator all at once~\cite{gehrmann-etal-2018-bottom}, it can happen that the generator arbitrarily stitches together text segments from unrelated sentences, yielding a summary that contains hallucinated content and fails to retain the meaning of the original document~\cite{falke-etal-2019-ranking,lebanoff-etal-2019-analyzing,kryscinski-etal-2019-neural}.

\begin{figure*}
\centering
\includegraphics[width=\textwidth]{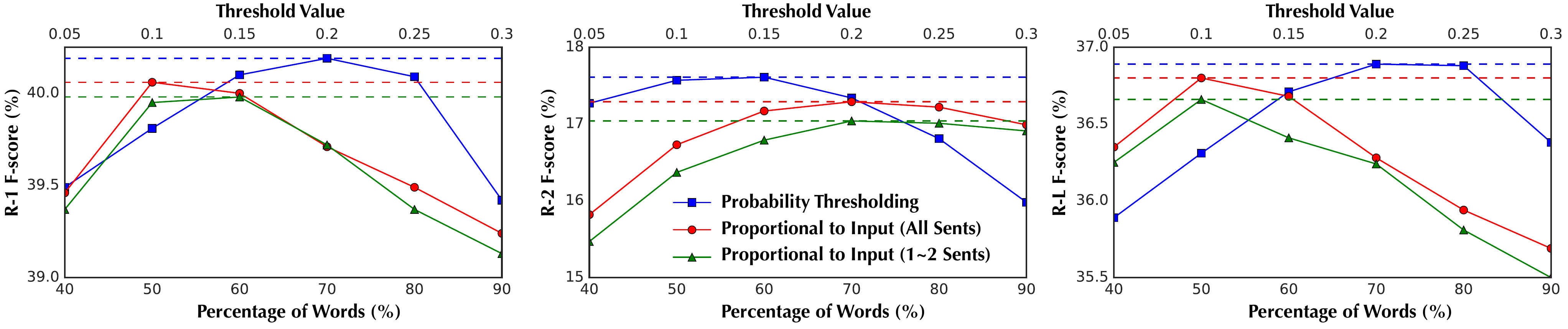}
\caption{Comparison of various highlighting strategies. Thresholding obtains the best performance.}
\label{fig:highlight_strategies}
\vspace{-0.1in}
\end{figure*}

We expect a sentence singleton or pair to be selected from the document if it contains salient content.
Moreover, a pair of sentences should contain content that is compatible with each other.
Given a sentence or pair of sentences from the document, our model predicts whether it is a valid instance to be compressed or merged to form a summary sentence. 
We follow \cite{lebanoff-etal-2019-scoring} to use BERT \cite{devlin2018bert} to perform the classification. 
BERT is a natural choice since it takes one or two sentences and generates a classification prediction.
It treats an input singleton or pair of sentences as a sequence of tokens. 
The tokens are fed to a series of Transformer block layers, consisting of multi-head self-attention modules. 
The first Transformer layer creates a contextual representation for each token, and each successive layer further refines those representations. 
An additional \textsf{\footnotesize[CLS]} token is added to contain the sentence representation. 
BERT is fine-tuned for our task by adding an output layer on top of the final layer representation $\mathbf{h}_{\scriptsize\mbox{[CLS]}}^L$ for sequence $s$, as seen in Eq.~(\ref{eq:p_sent}). 
\begin{align*}
p_{\scriptsize\mbox{sent}}(s) = \sigma(\mathbf{u}^\top \mathbf{h}_{\scriptsize\mbox{[CLS]}}^L)
\numberthis\label{eq:p_sent}
\end{align*}
where $\mathbf{u}$ is a vector of weights and $\sigma$ is the sigmoid function. The model predicts $p_{\scriptsize\mbox{sent}}$ -- whether the sentence singleton or pair is an appropriate one based on the \textsf{\footnotesize[CLS]} token representation.
We describe the training data for this task in \S\ref{sec:results}.

\vspace{0.08in}
\noindent\textbf{Fine-Grained Content Selection}\quad
It is interesting to note that the previous architecture can be naturally extended to perform fine-grained content selection by highlighting important words of sentences.
When two sentences are selected to generate a fusion sentence, it is desirable to identify segments of text from these sentences that are potentially compatible with each other.
The coarse-to-fine method allows us to examine the intermediate results and compare them with ground-truth.  
Concretely, we add a classification layer to the final layer representation $\mathbf{h}_i^L$ for each token $w_i$ (Eq.~(\ref{eq:p_highlight})).
The per-target-word loss is then interpolated with instance prediction (one or two sentences) loss using a coefficient $\lambda$.
Such a multi-task learning objective has been shown to improve performance on a number of tasks~\cite{guo-etal-2019-autosem}.
\begin{align*}
p_{\scriptsize\mbox{highlight}}(w_i) = \sigma(\mathbf{v}^\top \mathbf{h}_i^L)
\numberthis\label{eq:p_highlight}
\end{align*}
where $\mathbf{v}$ is a vector of weights and $\sigma$ is the sigmoid function.
The model predicts $p_{\scriptsize\mbox{highlight}}$ for each token -- whether the token should be included in the output fusion, calculated based on the given token's representation.

\vspace{0.08in}
\noindent\textbf{Information Fusion}\quad
Given one or two sentences taken from a document and their fine-grained highlights, we proceed by describing a fusion process that generates a summary sentence from the selected content.
Our model employs an encoder-decoder architecture based on pointer-generator networks that has shown strong performance on its own and with adaptations~\cite{see_get_2017,gehrmann-etal-2018-bottom}. 
We feed the sentence singleton or pair to the encoder along with highlights derived by the fine-grained content selector, the latter come in the form of binary tags. 
The tags are transformed to a ``\emph{highlight-on}'' embedding for each token if it is chosen by the content selector, and a ``\emph{highlight-off}'' embedding for each token not chosen. 
The highlight-on/off embeddings are added to token embeddings in an element-wise manner; both highlight and token embeddings are learned.
An illustration is shown in Figure~\ref{fig:model}. 

Highlights provide a valuable intermediate representation suitable for shallow abstraction.
Our approach thus provides an alternative to methods that use more sophisticated representations such as syntactic/semantic graphs~\cite{filippova-strube-2008-sentence,banarescu-etal-2013-abstract,liu-etal-2015-toward}.
It is more straightforward to incorporate highlights into an encoder-decoder fusion model, and obtaining highlights through sequence tagging can be potentially adapted to new domains.

\begin{table*}
\setlength{\tabcolsep}{4pt}
\renewcommand{\arraystretch}{1.1}
\begin{small}
\begin{minipage}[b]{0.47\hsize}
\centering
\begin{tabular}{lrrr}
\textbf{System} & \textbf{R-1} & \,\,\,\textbf{R-2} & \textbf{R-L} \\
\toprule
SumBasic{\scriptsize~\cite{vanderwende_beyond_2007}} & 34.11 & 11.13 & 31.14\\
LexRank{\scriptsize~\cite{erkan_lexrank:_2004}} & 35.34 & 13.31 & 31.93 \\
Pointer-Generator{\scriptsize~\cite{see_get_2017}} & 39.53 & 17.28 & 36.38\\
FastAbsSum{\scriptsize~\cite{chen-bansal-2018-fast}} & 40.88 & 17.80 & \textbf{38.54}\\
BERT-Extr{\scriptsize~\cite{lebanoff-etal-2019-scoring}} & 41.13 & \textbf{18.68} & 37.75\\
BottomUp{\scriptsize~\cite{gehrmann-etal-2018-bottom}} & \textbf{41.22} & \textbf{18.68} & 38.34\\
\midrule
BERT-Abs{\scriptsize~\cite{lebanoff-etal-2019-scoring}} & 37.15 & 15.22 & 34.60\\
Cascade-Fusion (Ours) & 40.10 & 17.61 & \textbf{36.71}\\
Cascade-Tag (Ours) & \textbf{40.24} & \textbf{18.33} & 36.14\\
\midrule
GT-Sent + Sys-Tag & 50.40 & 27.74 & 46.25\\
GT-Sent + Sys-Tag + Fusion & 51.33 & 28.08 & 47.50\\
GT-Sent + GT-Tag & \textbf{74.80} & 48.21 & \textbf{67.40}\\
GT-Sent + GT-Tag + Fusion & 72.70 & \textbf{48.33} & 67.06\\
\bottomrule
\end{tabular}

\end{minipage}
\hfill
\begin{minipage}[b]{0.55\hsize}
\centering

\begin{scriptsize}

\begin{tabular}{p{3in}}
\toprule

(\textsc{System Sents}) \textsf{\hlc[d]{A Duke student has admitted to hanging a noose made of rope from a tree near a student union, university} officials said Thursday. \hlc[l]{The student was identified during an investigation by campus police and the office of student affairs and admitted to placing the noose on the tree early Wednesday, the} university said.}\\[0.5em]

(\textsc{Cascade-Fusion}) \textsf{\hlc[d]{A Duke student} \hlc[l]{was identified during an investigation by campus police and the office of student affairs and admitted to placing the noose on the tree early Wednesday.}}\\

\midrule
    
(\textsc{GT Sents}) \textsf{In a news release, it said the \hlc[d]{student was no longer on campus and will face student conduct review.} \hlc[l]{Duke University is} a private college with about 15,000 \hlc[l]{students} in Durham, North Carolina.}\\[0.5em]

(\textsc{GT Sents + Fusion}) \textsf{\hlc[l]{Duke University} \hlc[d]{student was no longer on campus and will face student conduct review.}}\\[0.6em]

(\textsc{Reference}) \textsf{\hlc[d]{Student is no longer on} \hlc[l]{Duke University} \hlc[d]{campus and will face} disciplinary \hlc[d]{review.}}\\
    
\bottomrule

\end{tabular}
\end{scriptsize}

\end{minipage}
\end{small}
\caption{(\textsc{Left}) Summarization results on CNN/DM test set. Our cascade approach performs comparable to strong extractive and abstractive baselines; oracle models using ground-truth sentences and segment highlights perform the best. 
(\textsc{Right}) Example source sentences and their fusions. Dark highlighting is content taken from the first sentence, and light highlighting comes from the second. Our \textit{Cascade-Fusion} approach effectively performs entity replacement by replacing ``student'' in the second sentence with ``a Duke student'' from the first sentence.}
\label{tab:results-output}
\vspace{-0.1in}
\end{table*}

\section{Experimental Results}
\label{sec:results}

\vspace{0.08in}
\noindent\textbf{Data and Annotation}\quad
To enable direct comparison with end-to-end systems, we conduct experiments on the widely used CNN/DM dataset~\cite{see_get_2017} to report results of our cascade approach.
We use the procedure described in Lebanoff et al.~\shortcite{lebanoff-etal-2019-scoring} to create training instances for the sentence selector and fine-grained content selector. 
Our training data contains 1,053,993 instances; every instance contains one or two candidate sentences. 
It is a positive instance if a ground-truth summary sentence can be formed by compressing or merging sentences of the instance, negative otherwise.
For positive instances, we highlight all lemmatized unigrams appearing in the summary, excluding punctuation. We further add smoothing to the labels by highlighting single words that connect two highlighted phrases and by dehighlighting isolated stopwords. 
At test time, four highest-scored instances are selected per document; their important segments are highlighted by content selector then passed to the fusion step to produce a summary sentence each.
The hyperparameter $\lambda$ for weighing the per-target-word loss is set to 0.2 and highlighting threshold value is 0.15.
The model hyperparameters are tuned on the validation split.

\vspace{0.08in}
\noindent\textbf{Summarization Results}\quad
We show experimental results on the standard test set and evaluated by ROUGE metrics~\cite{lin-2004-rouge} in Table \ref{tab:results-output}.
The performance of our cascade approaches, \emph{Cascade-Fusion} and \emph{Cascade-Tag}, is comparable to or outranks a number of extractive and abstractive baselines.
Particularly, \emph{Cascade-Tag} does not use a fusion step (\S\ref{sec:framework}) and is the output of fine-grained content selection.
\emph{Cascade-Fusion} provides a direct comparison against BERT-Abs~\cite{lebanoff-etal-2019-scoring} that uses sentence selection and fusion but lacks a fine-grained content selector. 

Our results suggest that a coarse-to-fine content selection strategy remains necessary to guide the fusion model to produce informative sentences.
We observe that the addition of the fusion model has only a moderate impact on ROUGE scores, but the fusion process can reorder text segments to create true and grammatical sentences, as shown in Table~\ref{tab:results-output}. 
We analyze the performance of a number of oracle models that use ground-truth sentence selection (GT-Sent) and tagging (GT-Tag).
When given ground-truth sentences as input, our cascade models achieve $\sim$10 points of improvement in all ROUGE metrics. When the models are also given ground-truth highlights, they achieve an additional 20 points of improvement. 
In a preliminary examination, we observe that not all highlights are included in the summary during fusion, indicating there is space for improvement.
These results show that cascade architectures have great potential to generate shallow abstracts and future emphasis may be placed on accurate content selection.

\vspace{0.08in}
\noindent\textbf{How much should we highlight?}\quad
It is important to quantify the amount of highlighting required for generating a summary sentence.
Highlighting too much or too little can be unhelpful.
We experiment with three methods to determine the appropriate amount of words to highlight.
\emph{Probability Thresholding} chooses a set threshold whereby all words that have a probability higher than the threshold are highlighted. 
When \emph{Proportional to Input} is used, the highest probability words are iteratively highlighted until a target rate is reached.
The amount of highlighting can be proportional to the total number of words per instance (one or two sentences) or per document, containing all sentences selected for the document.

We investigate the effect of varying the amount of highlighting in Figure~\ref{fig:highlight_strategies}.
Among the three methods, probability thresholding performs the best, as it gives more freedom to content selection.
If the model scores all of the words in sentences highly, then we should correspondingly highlight all of the words. If only very few words score highly, then we should only pick those few. 

Highlighting a certain percentage of words tend to perform less well.
On our dataset, a threshold value of 0.15--0.20 produces the best ROUGE scores. 
Interestingly, these thresholds end up highlighting 58--78\% of the words of each sentence. 
Compared to what the generator was trained on, which had a median of 31\% of each sentence highlighted, the system's rate of highlighting is higher. 
If the model's highlighting rate is set to be similar to that of the ground-truth, it yields much lower ROUGE scores (cf. threshold value of 0.3 in Figure~\ref{fig:highlight_strategies}).
This observation suggests that the amount of highlighting can be related to the effectiveness of content selector and it may be better to highlight more than less.

\section{Conclusion}

We present a cascade approach to neural abstractive summarization that separates content selection from surface realization.
Importantly, our approach makes use of text highlights as intermediate representation; they are derived from one or two sentences using a coarse-to-fine content selection strategy, then passed to a neural text generator to compose a summary sentence.
A successful cascade approach is expected to accurately select sentences and highlight an appropriate amount of text, both can be customized for domain-specific tasks.

\section*{Acknowledgments}

We are grateful to the anonymous reviewers for their comments and suggestions.
This research was supported in part by the National Science Foundation grant IIS-1909603.

\bibliography{logan_summ,anthology,others}
\bibliographystyle{acl_natbib}

\end{document}